\documentclass[runningheads]{llncs}
\usepackage[T1]{fontenc}
\usepackage{graphicx,verbatim}
\usepackage{multirow}
\usepackage{amssymb}
\usepackage{float}
\usepackage{hyperref}
\usepackage{booktabs}
\usepackage{url}
\usepackage{todonotes}
\usepackage{makecell}
\usepackage{pifont}

\begin{document}
\title{Landmark-free Assessment of Lower-limb Alignment with Implicit Neural Shape Functions from Knee Radiographs}
\titlerunning{Implicit-based Lower-limb Alignment Assessment from Knee Radiographs}
%
\author{
Zhisen Hu$^{*}$\inst{1} \and
Antti Kemppainen\inst{2,3} \and
David Johnson\inst{4,5,6} \and
Egor Panfilov\inst{2} \and
Huy Hoang Nguyen\inst{2} \and
Timothy Cootes\inst{1} \and
Claudia Lindner\inst{1} \and
Aleksei Tiulpin\inst{7}
}
\authorrunning{Z. Hu et al.}
%
\institute{Division of Informatics, Imaging and Data Sciences, The University of Manchester, United Kingdom \and Research Unit of Health Sciences and Technology, University of Oulu, Finland \and Medical Research Center Oulu, University of Oulu and Oulu University Hospital, Finland \and Department of Trauma and Orthopaedics, Stockport NHS Foundation Trust, Stepping Hill Hospital, United Kingdom \and School of Health and Society, University of Salford, United Kingdom \and School of Biological Sciences, The University of Manchester, United Kingdom \and Weill Cornell Medicine, Cornell University, United States \\
\email{zhisen.hu@postgrad.manchester.ac.uk}}

\maketitle              

\begingroup
\renewcommand{\thefootnote}{}  
\footnotetext[1]{*Corresponding Author}
\addtocounter{footnote}{-1}    
\endgroup

\begin{abstract}
Radiographic assessment of lower-limb alignment (LLA) is important for predicting joint health and surgical outcomes in total knee arthroplasty. Traditional measurement methods are manual and time-consuming, while recent machine learning approaches typically rely on locating a fixed set of anatomical landmarks. This dependence limits flexibility and may require re-annotation when clinical definitions change. To address this, we propose an automated workflow using Implicit Neural Shape Functions (INSF). Rather than relying on explicit landmark coordinates, we encode the anatomy into a compact latent space and regress clinical alignment measurements directly from these latent codes. This architecture allows for rapid extendability to new tasks without altering the backbone representation. We trained our method on an internal dataset of 566 knee radiographs, each annotated with the outline of the femur and tibia. We evaluated it on both an internal test dataset of 50 patients and a separate external set of 402 preoperative cases from the MRKR dataset. Manual clinical measurements are available for these data, and the MRKR measurements will be made publicly accessible. Performance was comparable to state-of-the-art landmark-based methods and manual agreement, while offering a flexible shape representation that can be extended to additional measurement tasks.


\end{abstract}
%
%
\section{Introduction}

Knee osteoarthritis (OA) is a common and significant health issue that heavily burdens healthcare systems~\cite{steinmetz2023global}. Total knee replacement (TKR) may be offered as treatment for end-stage knee OA. Nevertheless, TKR is invasive, involving prosthesis implantation at the knee joint, and around 10$\%$ of patients are dissatisfied following TKR~\cite{ozden2025what,defrance2023are}. Pre-operative and post-operative lower-limb alignment (LLA) affects the outcomes following TKR, with radiographs revealing anomalies such as deformities of the femur and tibia, as well as incorrect positioning of the implants~\cite{ritter2011effect,ritter2013preoperative}. Accurate assessment of LLA in radiographs is important for successful treatment outcomes and long-term joint health. Traditional LLA measurement methods are manual and time-consuming. Machine learning-based automated techniques have now been widely used in the medical imaging area~\cite{zhang2025large,lin2025vascular}, including orthopaedics~\cite{lindner2013accurate,tiulpin2018automatic}. Such automated methods for measuring LLA in knee radiographs are potentially clinically valuable for reducing costs and improving the efficiency of the knee OA treatment pathway. 

\begin{figure}[t]
    \centering

    \begin{minipage}[b]{0.21\textwidth}
        \centering
        \includegraphics[width=\linewidth]{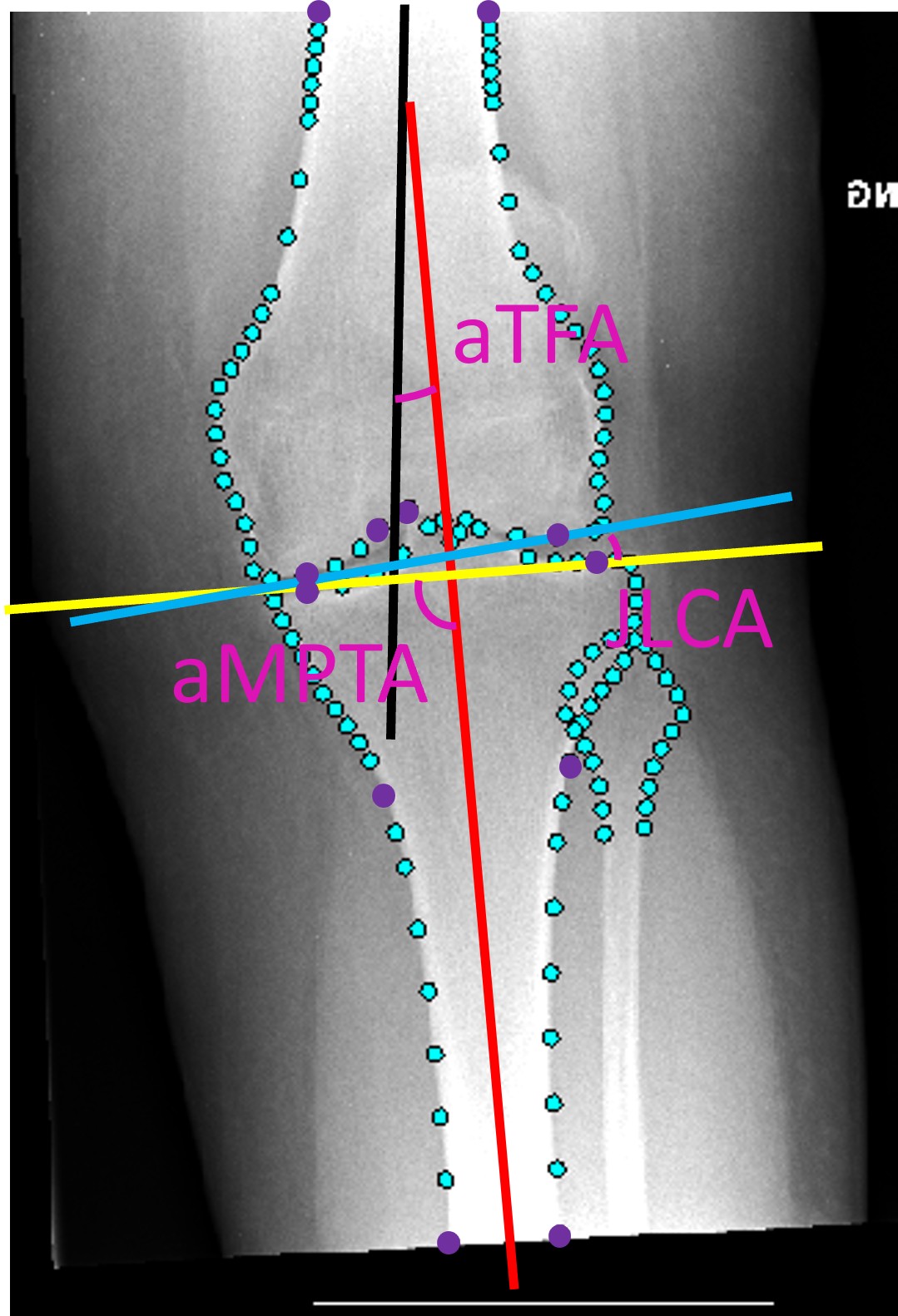}
        \\ (a)
    \end{minipage}
    \hfill
    \begin{minipage}[b]{0.65\textwidth}
        \centering
        \includegraphics[width=\linewidth]{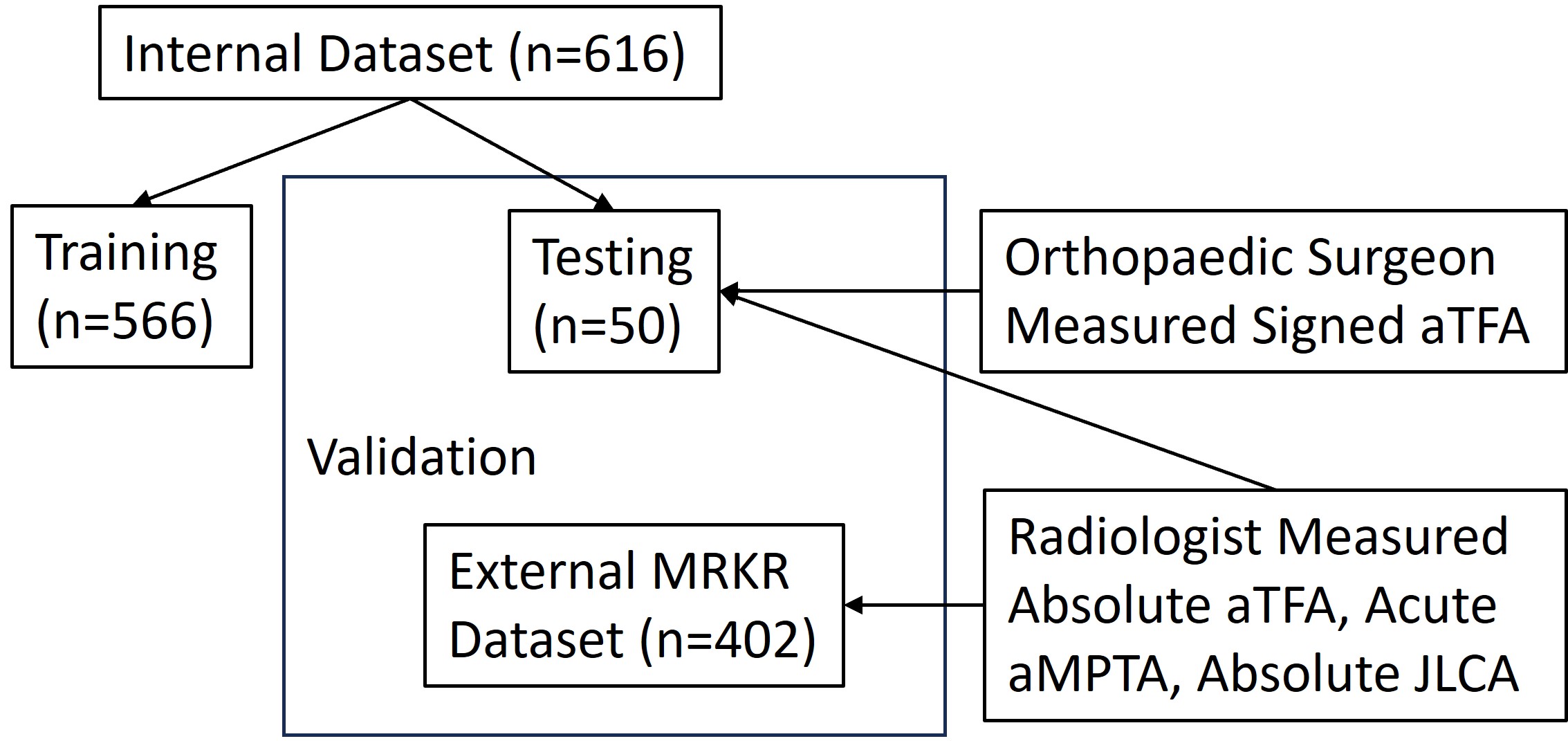}
        \\ (b) 
    \end{minipage}
    
    \caption{(a) Angles derived from landmarks. In landmark-based approaches, landmark positions (purple points) are used to fit several lines which form the angles. Black: anatomical femoral axis; Red: anatomical tibial axis; Blue: femoral joint line; Yellow: tibial joint line. aTFA is formed by the anatomical axes of femur (black) and tibia (red). aMPTA is formed by the anatomical tibial axis (red) and the tibial joint line (yellow), usually measured on the medial side. JLCA is formed by the femoral (blue) and tibial (yellow) joint lines. (b) Our study design. We trained our model with the internal training set and validated it using both internal and external testing sets.}
    \label{fig:intro}
\end{figure}

Recent machine learning approaches~\cite{cullen2025an,hu2025automated,hu2025deep} for measuring LLA primarily rely on point-based models to predict a predefined set of anatomical landmarks~\cite{lindner2013accurate,tiulpin2019kneel}, from which the angles representing LLA are subsequently computed. These landmarks must be densely and consistently annotated across images, which is both time-consuming and costly to perform. An example of generating angles such as anatomical tibio-femoral angle (aTFA), anatomical medial proximal angle (aMPTA), and joint line convergence angle (JLCA) from landmark positions is shown in Fig.~\ref{fig:intro}a. While this strategy enables direct computation of established clinical metrics, it inherently constrains the model to a discrete and predefined geometric representation of anatomy. 


Deep implicit shape representations~\cite{park2019deep} have been introduced to model shapes using a compact latent space and a signed distance function (SDF) auto-decoder. DISSM developed by Raju \textit{et al.}~\cite{raju2022deep} adopts this auto-decoder idea and proposes a complete workflow for modelling 3D liver and larynx shapes from CT images. In the orthopaedic field, for example, Pai \textit{et al.}~\cite{pai2025neural} employed neural shape models to quantify bone shape parameters from knee MRI. 

In this study, we leverage Implicit Neural Shape Functions (INSF) to model bone morphology from X-ray images using a SIREN-based auto-decoder framework~\cite{sitzmann2020implicit}. Instead of relying on explicit anatomical landmarks, the proposed method encodes bone geometry into a compact latent space and directly regresses LLA measurements from these learned representations using a multilayer perceptron (MLP). This continuous shape representation captures global anatomical structure and enables flexible adaptation to new clinical tasks without requiring predefined landmark definitions. To model multiple anatomical structures, the implicit representation is structured with one output channel per bone, allowing independent yet coordinated encoding of femoral and tibial geometry. The method was validated on both internal and external datasets (Fig.~\ref{fig:intro}b). To our knowledge, this is the first application of INSF to knee radiographs and the first use of implicit neural representations for automated LLA assessment.

\section{Methodology}

\subsection{Overview}

\begin{figure}[t]
    \centering
    \includegraphics[width=\linewidth]{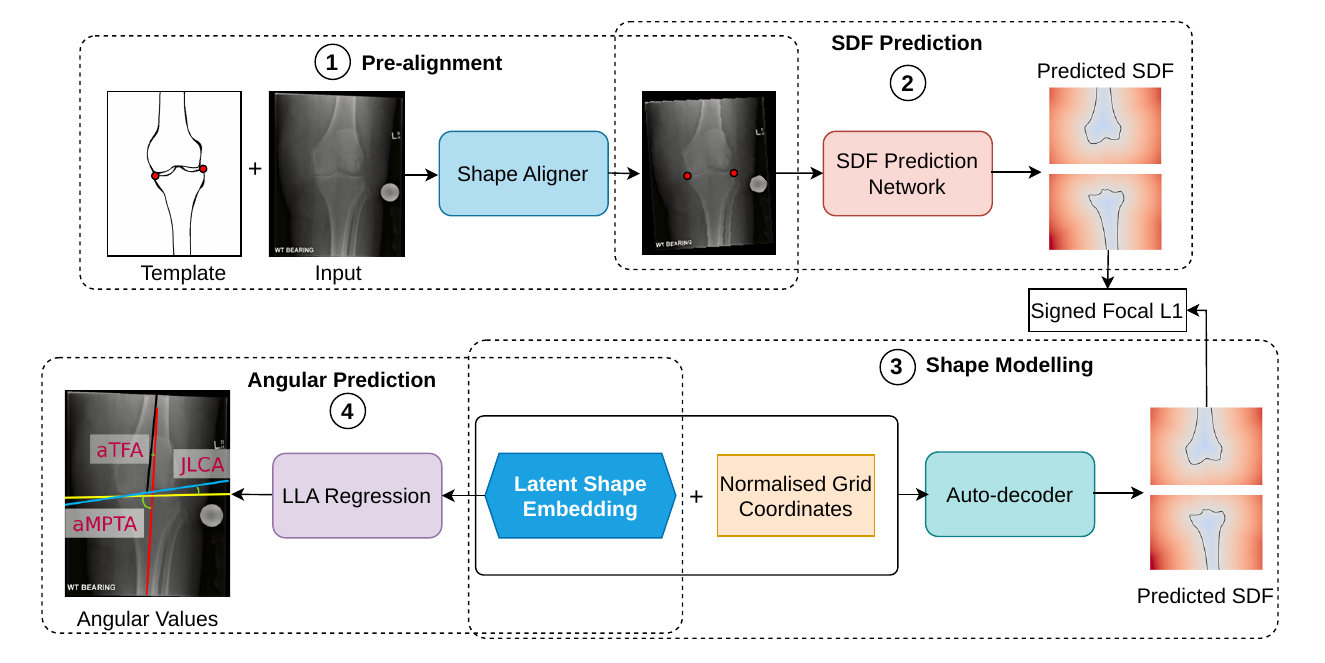}
    \caption{Overview of test-time shape embedding optimisation and LLA regression. Only latent shape embedding is optimised during test time.}
    \label{fig:overview}
\end{figure}

The overall inference workflow of our automated LLA assessment framework is illustrated in Fig.~\ref{fig:overview}. \ding{172} We first align all the shapes using two reference points at the corners of the tibial plateau. \ding{173} Then, we predict the SDFs from the aligned images. The predicted SDF contains two channels, with separate signed distance maps for the outlines of femur and tibia. \ding{174} Each testing sample is then directly fitted to the learned latent shape space via test-time optimisation. \ding{175} Finally, we directly regress the angular measurements with a compact 4-layer neural network (MLP). 

\subsection{Data}

\subsubsection{Image Data.}
Our internal image data consists of anonymised standard anteroposterior (AP) unilateral knee radiographs from patients undergoing TKR. To standardise anatomical orientation, all right knee images were horizontally flipped to appear as left knees. The radiographs were retrospectively collected from Stockport NHS Foundation Trust (approved by the Health Research Authority, IRAS 244130). This internal dataset includes 566 pre-operative images for training and 50 patients for testing. 

Our external dataset includes 402 knees, comprising both bilateral and unilateral images randomly selected from the Emory Knee Radiograph (MRKR) Dataset~\cite{price2024emory}. To ensure consistency with the internal data, bilateral images were split, and all right knee radiographs were flipped to maintain uniform orientation.

\subsubsection {LLA Angular Data.}

For training, ground-truth angles were derived from manually annotated landmarks. For internal testing, aTFA was independently measured by an orthopaedic surgeon and a radiologist. Each clinician performed two measurements 7–10 days apart, with the second blinded to the first; the mean of the two measurements was used as the final ground truth. The external evaluation included 402 cases from the Emory Knee Radiograph (MRKR) Dataset \cite{price2024emory}. Ground-truth measurements were obtained by the same radiologist. The radiologist recorded absolute aTFA and JLCA and measured the acute aMPTA.

\subsection{CNN-based SDF Prediction}
Following~\cite{hu2025automated,hu2025deep}, we first trained an Hourglass-based~\cite{newell2016stacked,tiulpin2019kneel} CNN to detect two reference points from the image for aligning all images into the canonical space. Subsequently, a U-Net~\cite{ronneberger2015unet} was trained to predict the canonical SDFs from the aligned knee images. 

As our dataset consists of knee radiographs centered on the knee joint, portions of the femoral or tibial shafts are occasionally truncated before reaching the image boundaries. It has been shown that incorporating a greater extent of shaft anatomy in the model improves the accuracy of aTFA measurements~\cite{cullen2025an}. To preserve anatomically consistent bone geometry and incorporate more shaft anatomy during training, we manually extended truncated shaft contours toward the image margins along estimated shaft tangents. This preprocessing step enables the U-Net model to learn consistent shaft geometry and to infer anatomically plausible shaft extensions toward the image boundaries at test time. 

\subsection{Implicit Shape Representation Learning}

We adopt SIREN~\cite{sitzmann2020implicit} as an auto-decoder to learn compact latent shape representations and to model bone geometries using SDFs, which output the distance from a queried spatial coordinate to the shape surface:

\begin{equation}
f_\theta(x,z) = d : x \in \mathbb{R}^2,\; d \in [-1,1]
\end{equation}

\noindent where $\theta$ is the network weights, $z$ is the latent shape code, $x$ is the input 2D normalised coordinates, the signed distance value $d$ is negative inside the shape and positive outside, with the surface defined by the zero level set ($d=0$). 

Following the auto-decoder paradigm in~\cite{park2019deep,raju2022deep}, we construct $K$ aligned canonical SDF training samples, $\{\mathcal{X}_k\}_{k=1}^{K}$, and associate each sample with a corresponding latent vector, $\{z_k\}_{k=1}^{K}$. The MLP decoder takes both spatial coordinates and the latent vector as input. During training, the network parameters and the latent shape embedding are optimised jointly. To better capture high-frequency geometric details, such as highly curved surface, we employ the Signed Focal L1 loss proposed in~\cite{dang2024singr}. The overall optimisation objective is:

\begin{equation}
\arg\min_{\theta, z}
\sum_{k=1}^{K}
\left(
\sum_{i=1}^{|{\mathcal{X}}_k|}
\frac{1}{|\Omega|}
\sum_{i\in\Omega}
\left| S_i - P_i,_k \right|
\frac{
\left| S_i - P_i,_k \right|^{\gamma}
\;\mathbb{I}\!\left(S_i P_i,_k \ge 0\right)
}{
\max\!\left(|S_i|,\left|P_i,_k\right|\right)
+ \epsilon
}
\right)
\end{equation}

\noindent where $P_i,_k$ includes the use of the tanh function to convert the network output into the range $[-1,1]$ ($P_i,_k=\tanh\!\big(f_\theta(x_i,z_k)\big)$), $S$ is the ground-truth SDF value, $\epsilon$ is a positive constant to avoid numerical issues, $\gamma$ is a positive hyperparameter, and $\mathbb{I}(\cdot)$ is the indicator function. 

\subsection{Test-time Optimisation and LLA Regression}

Prior work typically fits each sample to the implicit shape model by training an additional network to predict the latent code~\cite{raju2022deep}. We use a simpler approach. We use a U-Net to predict the SDF for each image. We then perform an optimisation to find the latent vector, z, which minimises the focal loss between the generated SDF and that from the U-Net.

Unlike traditional methods for LLA assessment, which explicitly compute geometric measurements from localised anatomical landmarks, we directly regress the angular values that quantify LLA using a simple MLP. The MLP takes the latent shape codes as input and outputs the corresponding angle measurements.

\section{Results}

\subsubsection{Shape Modelling.}
\begin{figure}[ht]
    \centering
    
    \begin{minipage}[b]{0.32\textwidth}
        \centering
        \includegraphics[height=3cm,keepaspectratio]{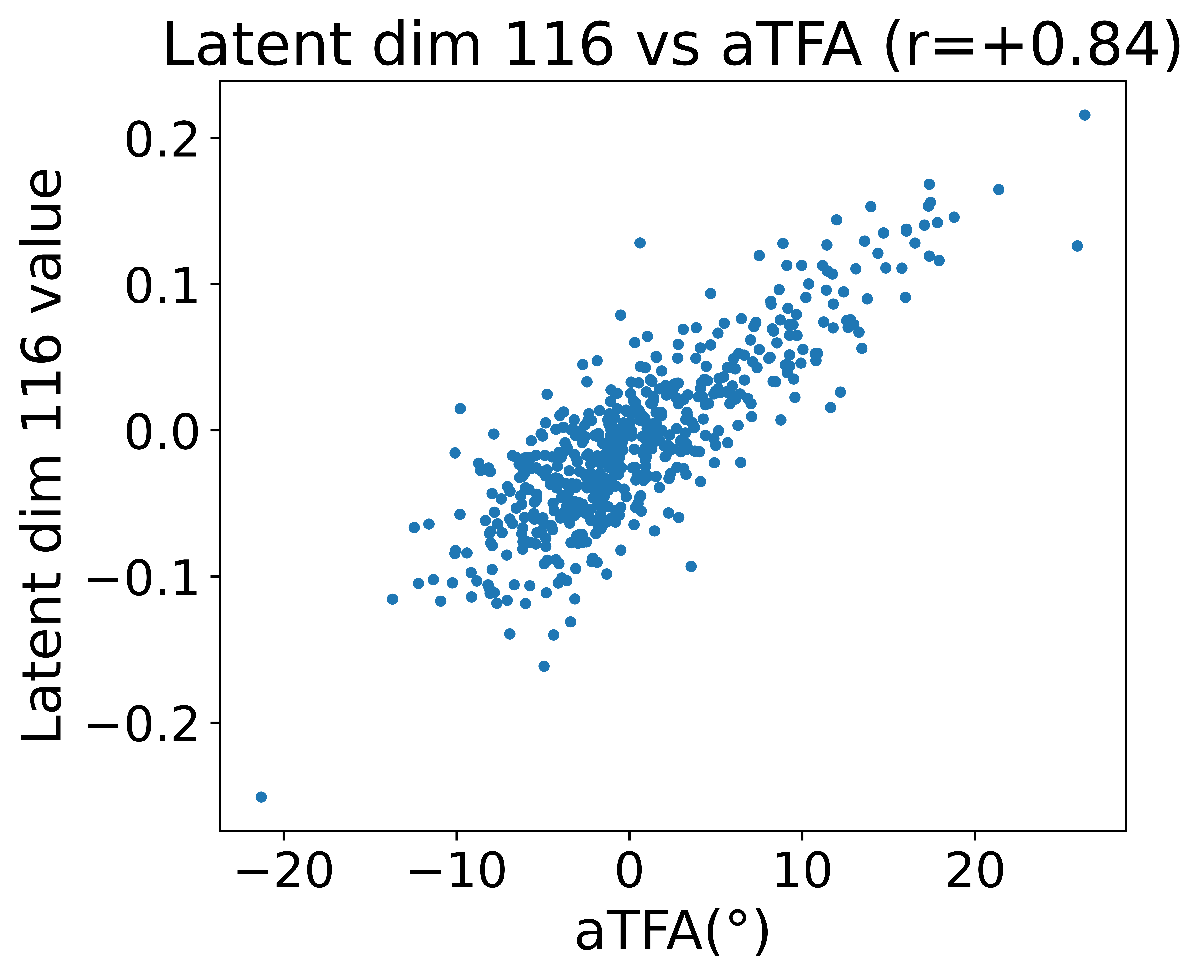}
        \\ (a) 
    \end{minipage}
    \hfill
    \begin{minipage}[b]{0.32\textwidth}
        \centering
        \includegraphics[height=3cm,keepaspectratio]{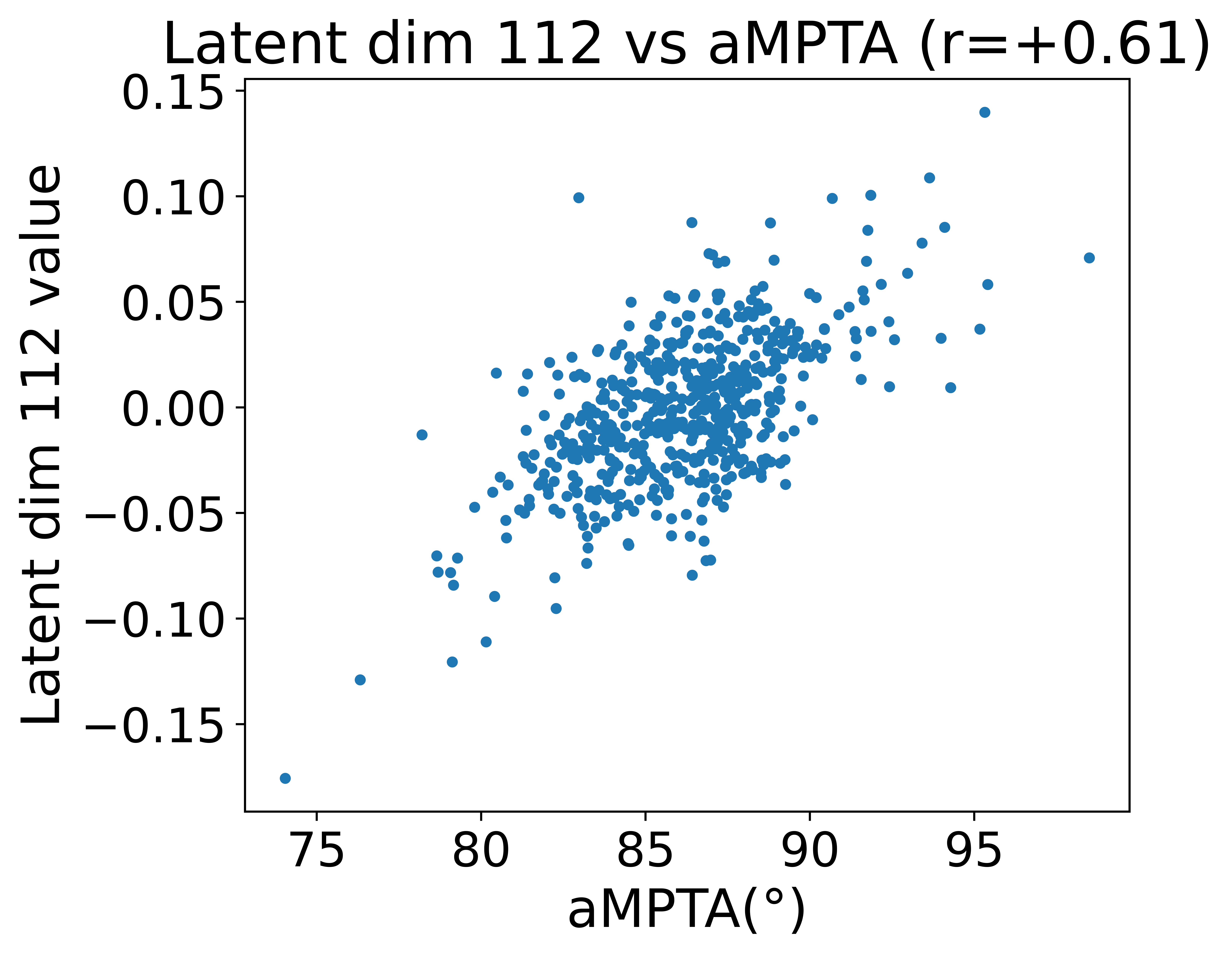}
        \\ (b) 
    \end{minipage}
    \hfill
    \begin{minipage}[b]{0.32\textwidth}
        \centering
        \includegraphics[height=3cm,keepaspectratio]{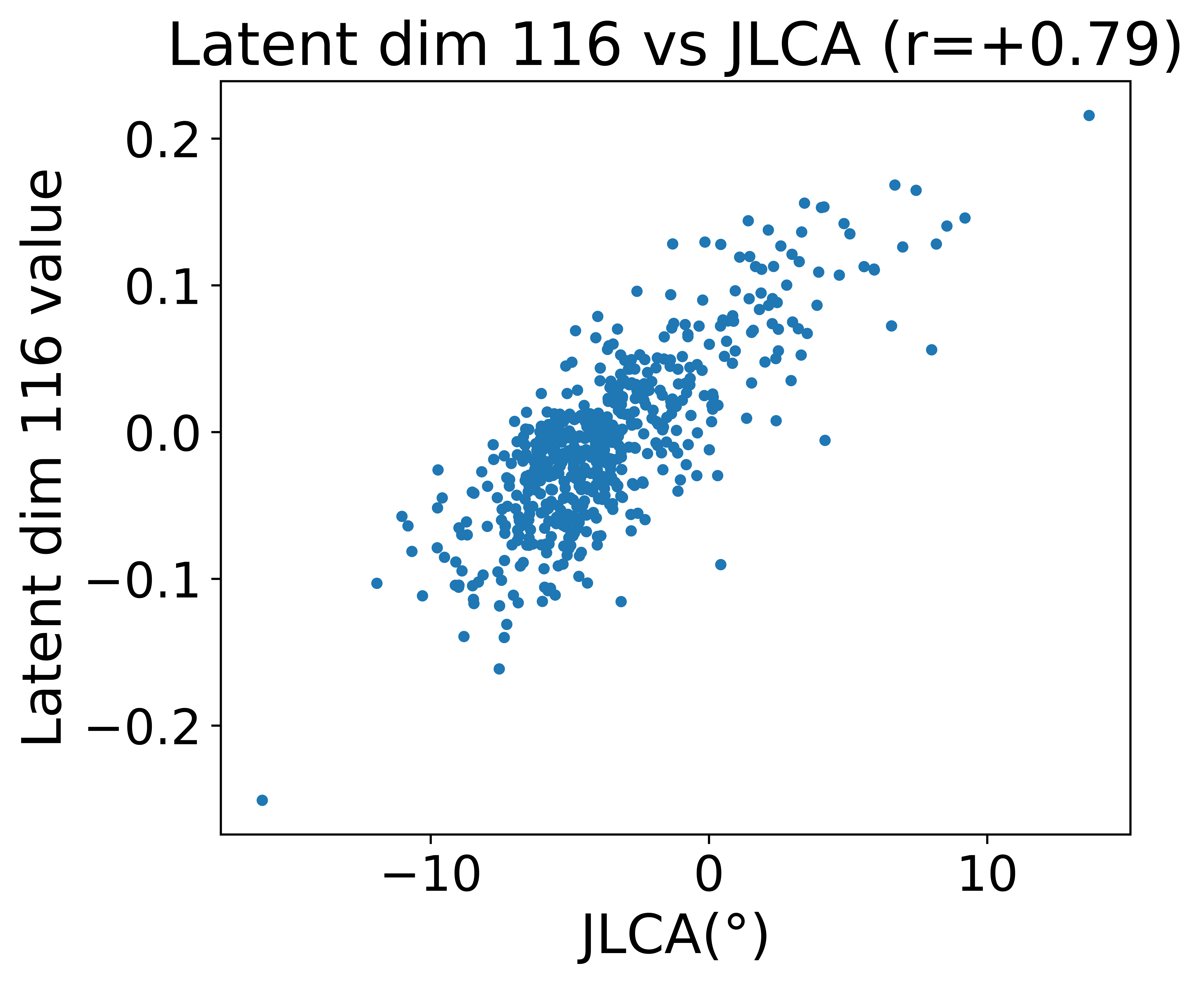}
        \\ (c) 
    \end{minipage}
    

    \begin{minipage}[b]{0.49\textwidth}
        \centering
        \includegraphics[width=0.33\linewidth]{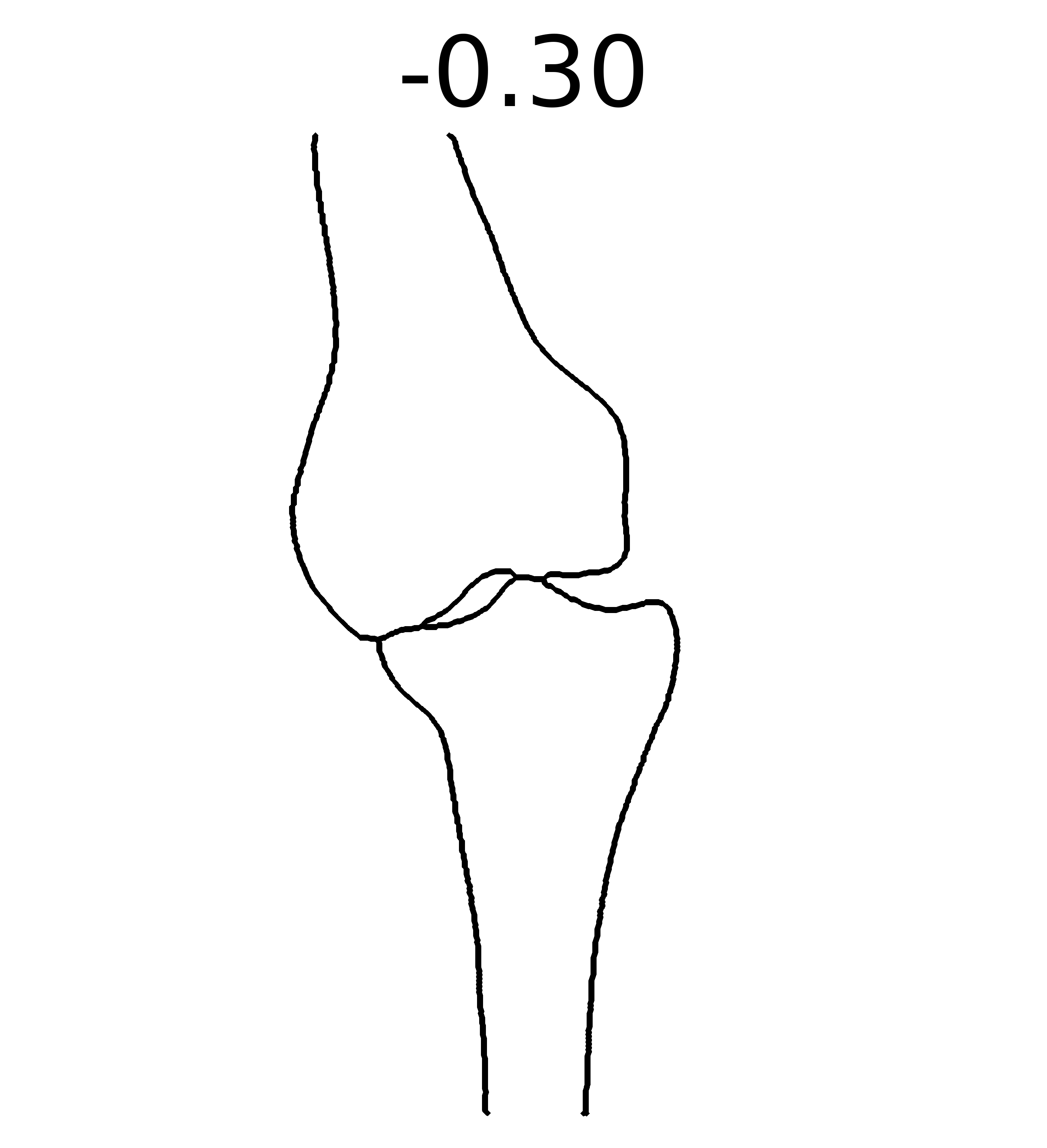}\hfill
        \includegraphics[width=0.33\linewidth]{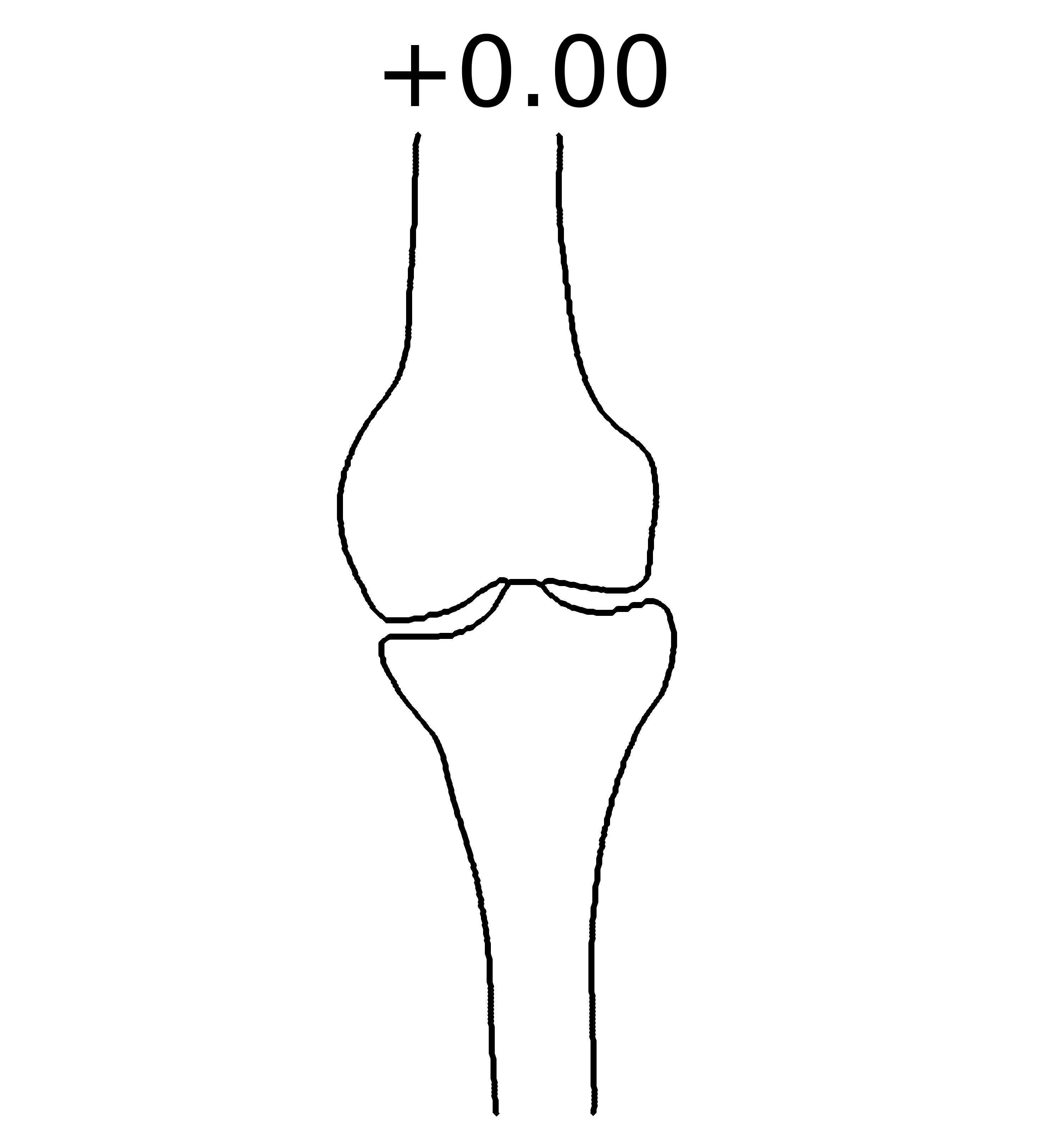}\hfill
        \includegraphics[width=0.33\linewidth]{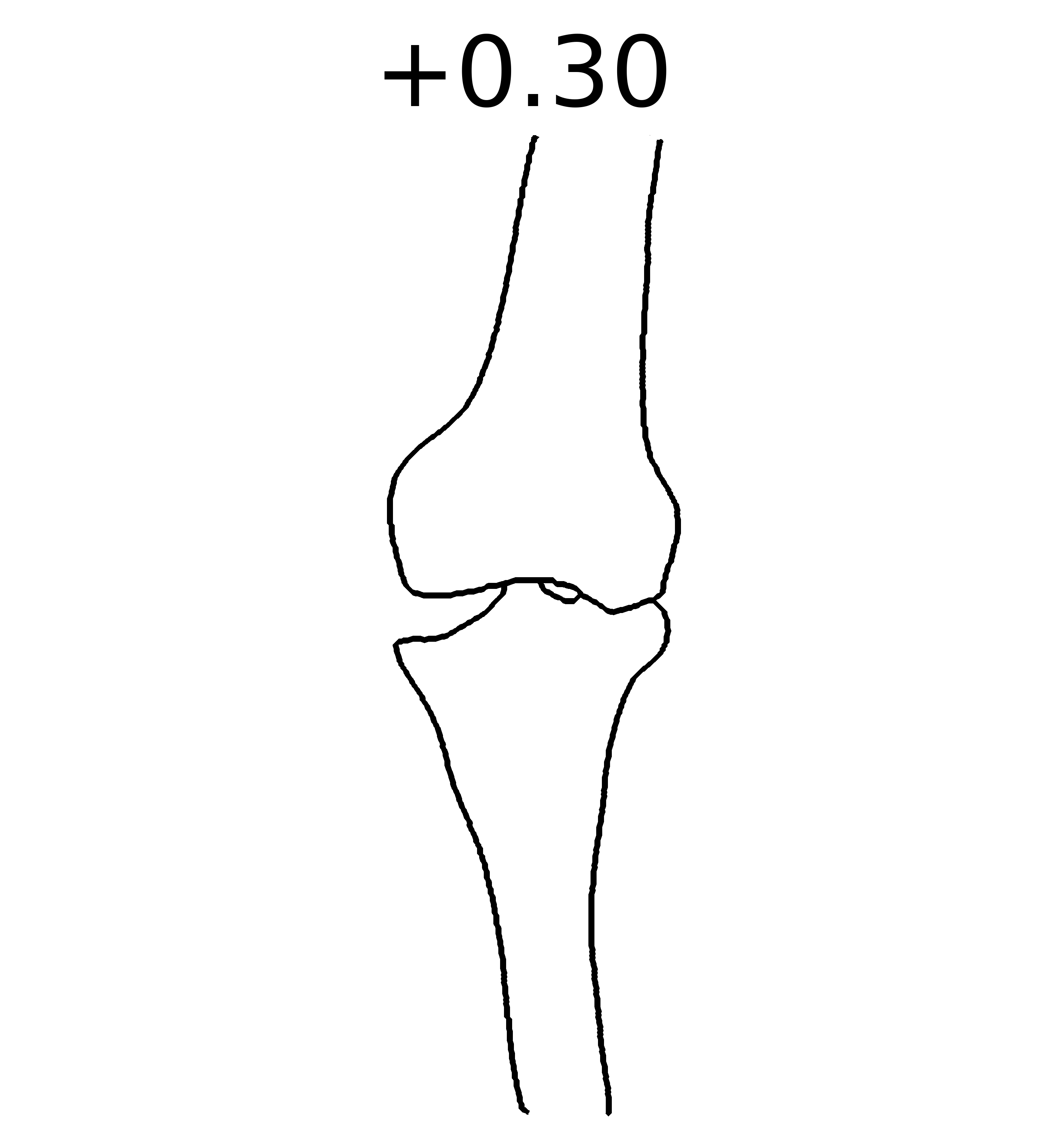}
        \\ (d) Shape Variation on Latent Dimension 116
    \end{minipage}
    \hfill
    \begin{minipage}[b]{0.49\textwidth}
        \centering
        \includegraphics[width=0.33\linewidth]{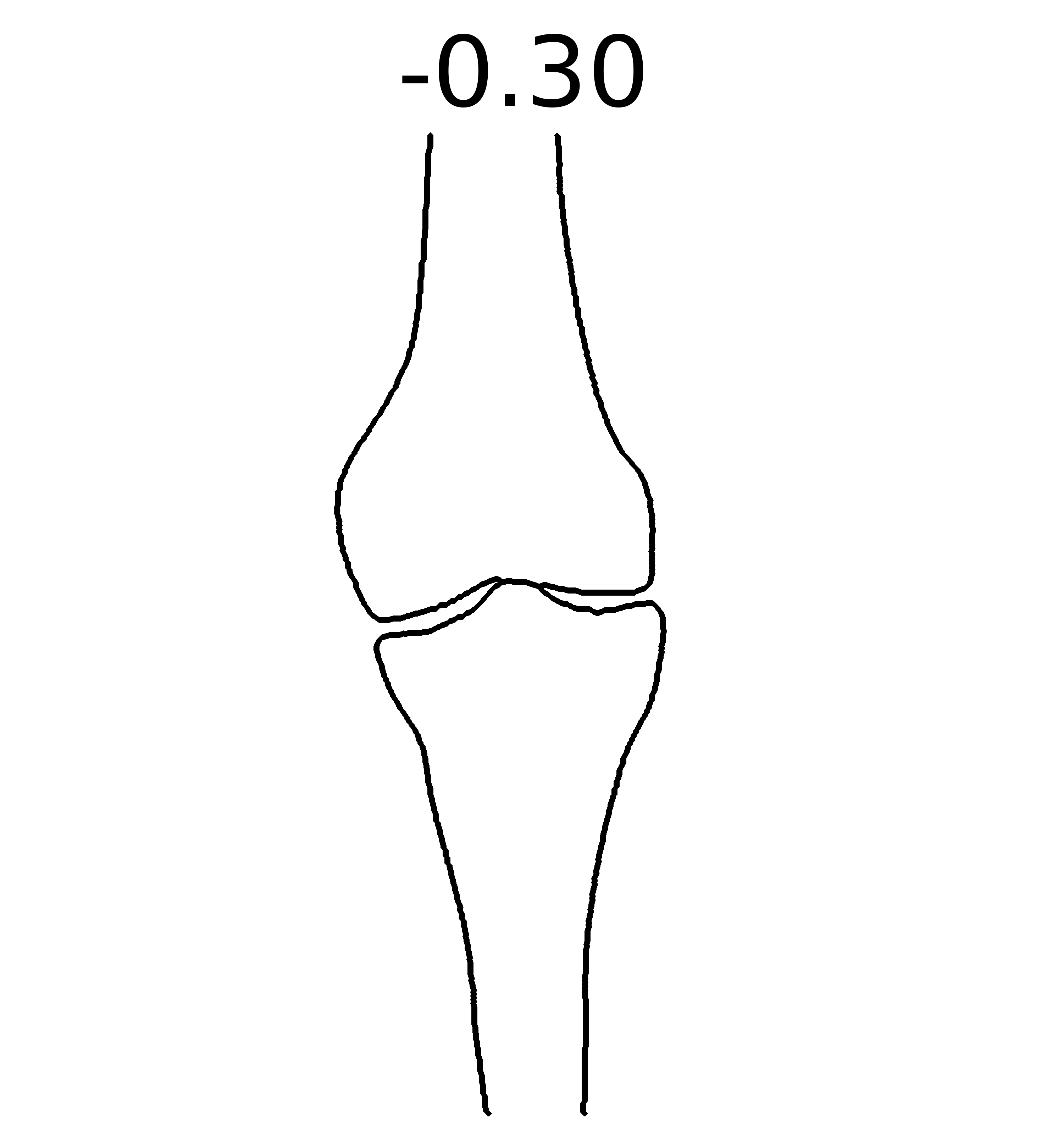}\hfill
        \includegraphics[width=0.33\linewidth]{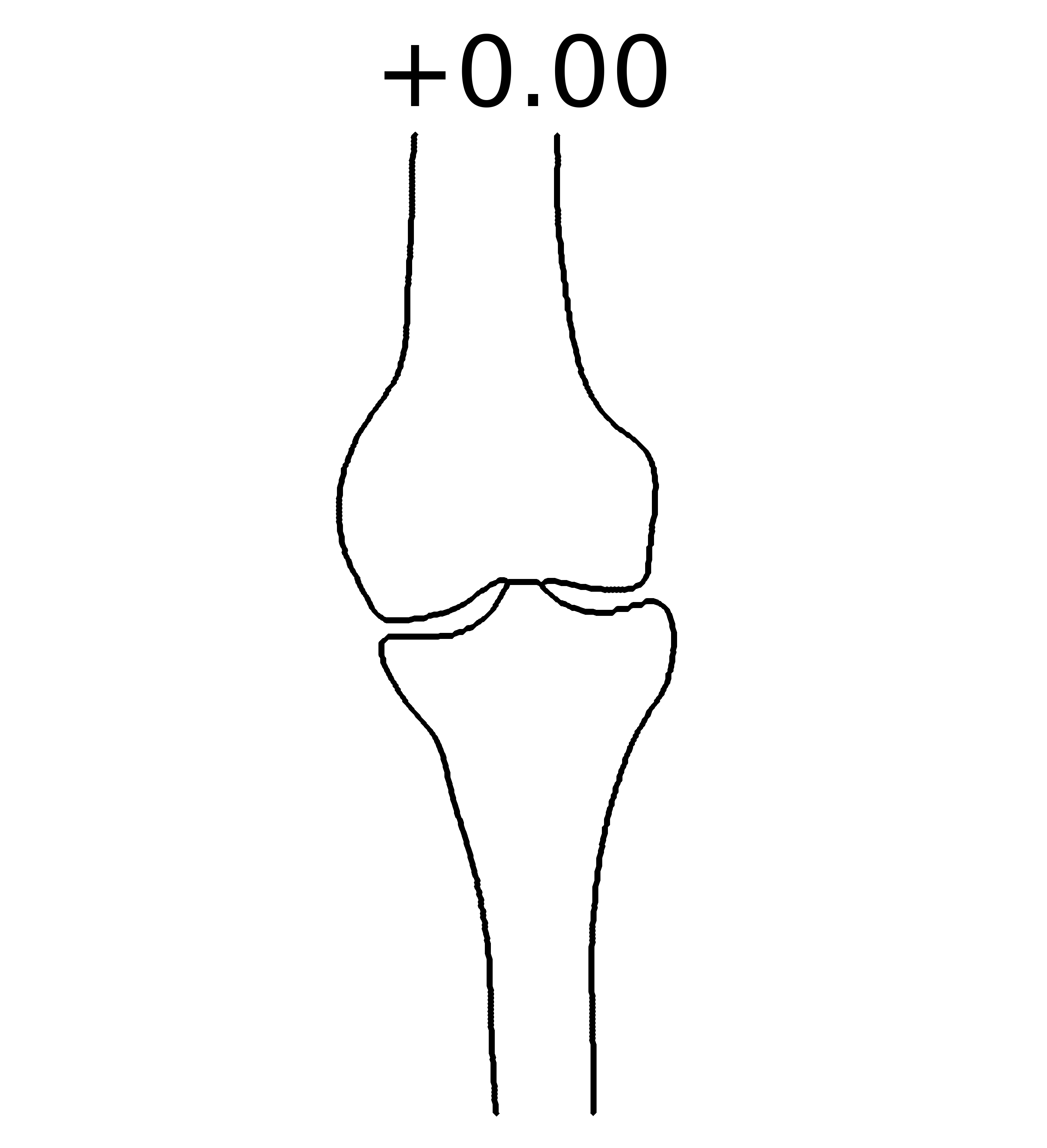}\hfill
        \includegraphics[width=0.33\linewidth]{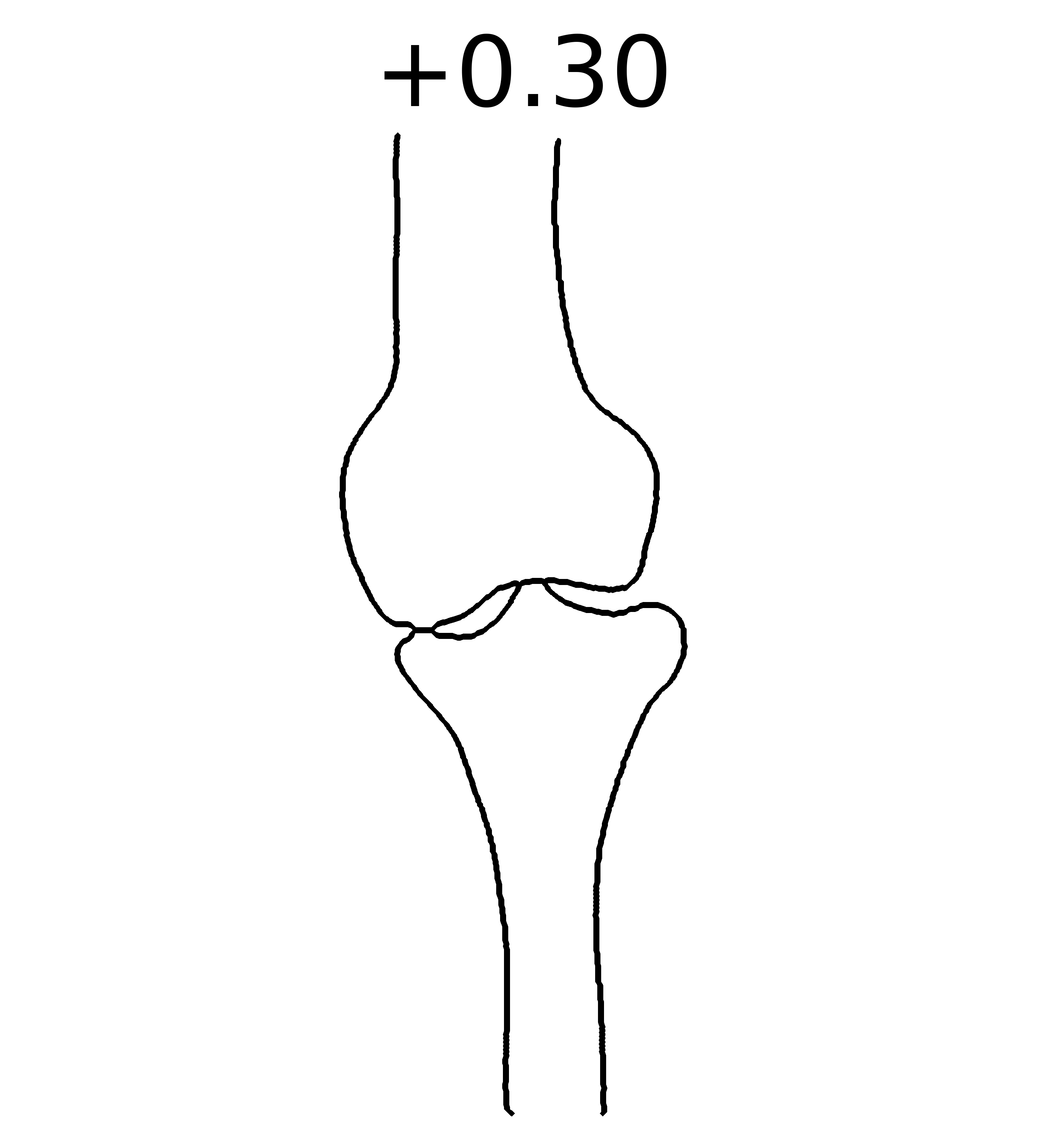}
        \\(e) Shape Variation on Latent Dimension 112
    \end{minipage}
    
    \caption{(a-c) Correlation analysis between selected latent dimensions and angles, and (d-e) visualisation of shape variation on the most correlated dimensions.}
    \label{fig:latent_correlation}
\end{figure}

We constructed a latent shape space represented by 128-dimensional vectors. To investigate the relationship between each latent dimension and LLA, we computed the Pearson correlation coefficient ($|r|$) between the latent codes and the angular values in the training set. The dimensions were then ranked according to their absolute correlation coefficients. All three angles exhibited at least moderate correlations ($|r|>0.6$) with their respective most strongly associated latent dimensions. Notably, signed aTFA and JLCA demonstrated strong correlations ($|r|\approx0.8$) with their most correlated latent dimensions (Fig.~\ref{fig:latent_correlation}a and~\ref{fig:latent_correlation}c). Latent dimension 116 exhibited the strongest correlation with both aTFA and JLCA, whereas latent dimension 112 was most strongly associated with aMPTA. The latent dimensions are learned in an unsupervised manner and are not explicitly constrained to correspond to predefined anatomical variables. Consequently, the numerical indices (e.g., 116 and 112) have no inherent meaning and are not fixed: the identity and ordering of the dimensions can change across training runs. A given dimension becomes interpretable only through its observed statistical association with the angular measurements.

We further analysed how variations on the most strongly correlated dimension of each angle (116, 112) affected the shape contours, similar to visualisations in point-based shape models~\cite{cootes1995active,cootes2012robust}. We visualised the mean shape and systematically varied the corresponding latent parameter from –0.3 to 0.3 while keeping all other dimensions fixed (Fig.~\ref{fig:latent_correlation}d and~\ref{fig:latent_correlation}e). For latent dimension 116, increasing the parameter produced progressive rightward shifts of the femoral and tibial shafts, indicating that the learned shape model automatically captures pose-related information that influences aTFA and JLCA, thereby affecting LLA. 

\subsubsection{LLA Assessment.}
Intraclass correlation coefficient (ICC) and mean absolute difference (MAD) were used to evaluate the agreement between the measurements assessed in various ways. We first compared our approach with the methods proposed in~\cite{cullen2025an,hu2025deep}, as well as with the intra- and inter-rater agreements, using the aTFA measurements obtained by both clinicians on the internal dataset ($n=50$). When comparing automated methods with the ground-truth aTFA values measured by the orthopaedic surgeon, our method achieved comparable results ($MAD=1.2$\textdegree{}; $ICC=0.97$) with both~\cite{cullen2025an} ($MAD=1.2$\textdegree{}; $ICC=0.97$) and~\cite{hu2025deep} ($MAD=1.1$\textdegree{}; $ICC=0.97$). However, all automated methods performed marginally worse than the intra-rater agreement. For comparison with ground-truth aTFA values measured by the radiologist, automated methods were comparable to intra- or inter-rater agreements ($MAD\approx1.0$\textdegree{}; $ICC\approx0.95$). 

When stratified by shaft length, our auto–shaft extension approach outperformed the same method without shaft extension for aTFA measurement. Specifically, when compared with the orthopaedic surgeon’s measurements, the MAD decreased from $1.8$\textdegree{} to $1.2$\textdegree{}, and the ICC improved from $0.94$ to $0.97$. Similarly, when compared with the radiologist’s measurements, the MAD decreased from $1.6$\textdegree{} to $1.0$\textdegree{}, and the ICC increased from $0.87$ to $0.94$. These findings demonstrate that incorporating shaft extension enhances both measurement accuracy and agreement with expert assessments.

We additionally evaluated our method on a subset ($n=402$) of the external MRKR dataset~\cite{price2024emory}, comparing it with the approach in~\cite{hu2025deep} and manual intra-rater agreement. In addition to aTFA, we included aMPTA and JLCA in the generalisation experiment to assess extendability. On this dataset, our method achieved performance comparable to~\cite{hu2025deep} for aTFA, with no statistically significant difference observed. For the other angles, our method showed marginally lower numerical performance. Both automated approaches underperformed relative to manual intra-rater agreement.

The comparison of our method,~\cite{hu2025deep}, along with the intra-rater agreement, is presented in Table~\ref{tab:results}. Statistical testing was performed to determine whether the MAD exceeded $\pm1$\textdegree{}, using a two one-sided tests scheme with Bonferroni–Holm adjustment. All $p$-values were $<0.0001$ and are therefore not reported. Overall, the statistical analysis indicates that the evaluated methods perform equivalently within a $\pm1$\textdegree{} difference threshold.

\begin{table}[ht]
\centering
\caption{Agreement analysis: mean absolute difference (MAD; \textdegree{}) and intraclass correlation coefficient (ICC) for internal and external datasets. Statistical testing was done to verify if the MAD exceeds $\pm1$\textdegree{}. All $p$-values were $<0.0001$. Here, OS indicates orthopaedic surgeon and RA indicates board-certified radiologist. }
\label{tab:results}
\begin{tabular}{p{1.5cm}p{1cm}llll}
\toprule
\textbf{Dataset} & \textbf{\#} & \textbf{Angle} & \textbf{Method} & \textbf{MAD (\textdegree{})} & \textbf{ICC} \\
\midrule
Internal & 50 & aTFA & Ours vs.\ OS & $1.19_{\,(0.92,\,1.54)}$ & $0.97_{\,(0.94,\,0.98)}$ \\
& &  & Hu \textit{et al.}~\cite{hu2025deep} vs.\ OS & $1.09_{\,(0.84,\,1.40)}$ & $0.97_{\,(0.95,\,0.98)}$ \\ \cmidrule{4-6}
& &  & Ours vs.\ RA & $1.03_{\,(0.80,\,1.32)}$ & $0.94_{\,(0.90,\,0.96)}$ \\ 
& &  & Hu \textit{et al.}~\cite{hu2025deep} vs.\ RA & $0.95_{\,(0.74,\,1.22)}$ & $0.96_{\,(0.92,\,0.97)}$ \\ \cmidrule{4-6}
& &  & OS (intra-rater) & $0.89_{\,(0.69,\,1.14)}$ & $0.99_{\,(0.98,\,0.99)}$ \\
&  &  & RA (intra-rater) & $0.94_{\,(0.73,\,1.20)}$ & $0.95_{\,(0.92,\,0.97)}$ \\
&  &  & OS vs.\ RA (inter-rater) & $0.97_{\,(0.75,\,1.26)}$ & $0.95_{\,(0.90,\,0.97)}$ \\ \midrule
External & 402 &  aTFA & Ours & $1.08_{\,(0.98,\,1.19)}$ & $0.84_{\,(0.81,\,0.87)}$ \\
& &  & Hu \textit{et al.}~\cite{hu2025deep} & $0.91_{\,(0.83,\,1.01)}$ & $0.87_{\,(0.85,\,0.90)}$ \\
&  &  & RA (intra-rater) & $0.65_{\,(0.59,\,0.72)}$ & $0.93_{\,(0.92,\,0.94)}$ \\
\cmidrule{2-6}
& & aMPTA & Ours & $1.29_{\,(1.17,\,1.42)}$ & $0.58_{\,(0.30,\,0.74)}$ \\
& &  & Hu \textit{et al.}~\cite{hu2025deep} & $0.88_{\,(0.80,\,0.97)}$ & $0.72_{\,(0.64,\,0.78)}$ \\
& &  & RA (intra-rater)  & $0.77_{\,(0.70,\,0.85)}$ & $0.79_{\,(0.76,\,0.83)}$ \\
\cmidrule{2-6}
& & JLCA & Ours & $0.94_{\,(0.85,\,1.03)}$ & $0.67_{\,(0.61,\,0.72)}$ \\
& &  & Hu \textit{et al.}~\cite{hu2025deep} & $0.75_{\,(0.68,\,0.83)}$ & $0.80_{\,(0.72,\,0.86)}$ \\
& &  & RA (intra-rater)  & $0.49_{\,(0.45,\,0.54)}$ & $0.89_{\,(0.86,\,0.91)}$ \\
\bottomrule
\end{tabular}
\end{table}

\section{Discussions and Conclusions}

In this paper, we have shown that implicit neural representations capture variations associated with LLA. The proposed LLA assessment workflow achieved performance comparable to landmark-based methods and to manual intra- or inter-rater agreement on the internal dataset, but declined slightly on the external dataset, where it was slightly inferior to the landmark-based method and manual agreement.  This reduction in generalisability may reflect population differences. The internal dataset predominantly comprised White patients, whereas the external MRKR dataset~\cite{price2024emory} included approximately 40$\%$ White and 40$\%$ African American patients. In addition, the mean age of the internal dataset was approximately ten years higher than that of the MRKR cohort. These findings suggest that further refinement may improve robustness across diverse populations.

Performance differences were generally small between methods, and variability in clinician annotations and measurement protocols may have influenced the results. We observed no significant differences between automated approaches, as performance is inherently dependent on the underlying clinical annotations. Despite minor performance discrepancies, none of the observed effects was statistically or clinically significant. The methods were found equivalent to the ground truth measurements, as well as to each other ($p<0.0001$).

When it comes to the interpretability of the latent space, its correlations with aTFA or JLCA were strong, whereas the association with aMPTA was comparatively weaker. This may reflect the model’s more accurate representation of relative bone positioning rather than displacement of individual bones. 

A key limitation of this study is that we utilized the pre-processing step from~\cite{hu2025automated,tiulpin2019kneel} for shape pre-alignment out of convenience. This, however, can easily be incorporated within one UNet model. In general, our training targets were derived from landmark-based annotations, which may inherently favor landmark-driven approaches and constrain the evaluation of the full advantages offered by a landmark-free representation. Future work should focus on directly regressing measurements that cannot be readily derived from landmark positions. This would enable a more comprehensive assessment of the proposed approach and more convincingly demonstrate its extendability.

Nevertheless, our framework offers important advantages. By encoding global anatomical shape information into latent representations, it learns LLA-related features without predefined landmarks. This enables compact shape representation in latent vectors and facilitates extension to new clinical tasks. Such flexibility is particularly valuable when landmarks are difficult to detect or inconsistently defined, making the framework well suited to evolving clinical definitions where traditional point-based models are impractical. 

    



%
%
%
%

\end{document}